\documentclass[11pt]{article}

\usepackage[margin=1in]{geometry}

\usepackage{amsmath,amsfonts}

\usepackage{graphicx}

\usepackage{physics}

\usepackage{hyperref}

\usepackage{cite}

\usepackage{booktabs}

\usepackage{subcaption}

\title{SPUS: A Lightweight and Parameter-Efficient Foundation Model for PDEs}

\author{
Abu Bucker Siddik\textsuperscript{*1},\quad
Diane Oyen$^{1}$,\quad
Alexander Most $^{1}$,\quad
Michal Kucer $^{2}$,\quad
Ayan Biswas$^{1}$\quad
 \\
$^1$ Computing and Artificial Intelligence Division (CAI) \\
$^2$ Space Remote Sensing and Data Science \\
Los Alamos National Laboratory, Los Alamos, New Mexico, US, 87545 
}

\begin{document}
 
\maketitle
\author{\thanks{*Corresponding author: \texttt{siddik@lanl.gov}}}

\begin{abstract}
 We introduce Small PDE U-Net Solver (SPUS), a compact and efficient foundation model (FM) designed as a unified neural operator for solving a wide range of partial differential equations (PDEs). Unlike existing state-of-the-art PDE FMs—primarily based on large complex transformer architectures with high computational and parameter overhead—SPUS leverages a lightweight residual U-Net-based architecture that has been largely underexplored as a foundation model architecture in this domain. To enable effective learning in this minimalist framework, we utilize a simple yet powerful auto-regressive pretraining strategy which closely replicates the behavior of numerical solvers to learn the underlying physics. SPUS is pretrained on a diverse set of fluid dynamics PDEs and evaluated across 6 challenging unseen downstream PDEs spanning various physical systems. Experimental results demonstrate that SPUS using residual U-Net based architecture achieves state-of-the-art generalization on these downstream tasks while requiring significantly fewer parameters and minimal fine-tuning data, highlighting its potential as a highly parameter-efficient FM for solving diverse PDE systems.
\end{abstract}

\section{Introduction}
\label{Introduction}
Partial differential equations (PDEs) are fundamental mathematical tools for modeling a wide range of complex spatio-temporal phenomena in science and engineering, including fluid dynamics, electromagnetism, materials science, and climate systems \cite{neumann2012coupled, schaa2016pde, muller2014massively}. Traditional numerical solvers—such as finite difference and finite element methods—are widely used for PDE simulation but often come with high computational costs, especially when repeated simulations are required for varying coefficients or boundary conditions \cite{herde2024poseidon}. To address these limitations, deep learning–based approaches like the Fourier neural operator \cite{li2021fourier}, convolutional neural operator \cite{raonic2023convolutional}, and DeepONet \cite{lu2021learning} have been proposed. While these models have shown promising performance, they are typically designed for specific PDE families and require retraining when applied to new classes of governing equations, leading to significant computational overhead. 

Some simulation data is more computationally expensive to produce from numerical solvers than others; and so multiphysics PDE FMs take advantage of pretraining on large benchmark PDE data to finetune on limited PDE data from more expensive simulations. PDE FMs—including MPP \cite{mccabe2024multiple}, POSEIDON \cite{herde2024poseidon}, PROSE FD \cite{liu2024prose}, and DPOT \cite{hao2024dpot}—have emerged as a promising paradigm. These models aim to learn unified representations by incorporating multiple physical systems into a single framework, demonstrating the ability to generalize to unseen PDE families using limited data. However, current state-of-the-art FM approaches predominantly utilize transformer-based architectures with high parameter counts in the hundreds of millions, resulting in increased computational and data demands \cite{mccabe2024multiple, herde2024poseidon, hao2024dpot}. To overcome these limitations, in this work, we propose a efficient yet effective Small PDE U-Net Solver (SPUS), with an order of magnitude fewer parameters, for PDE foundation modeling. To the best of our knowledge, this is the first work to explore a residual U-Net architecture as an FM pretrained on a large and diverse PDE dataset, beyond single-family PDE prediction. U-Net has been shown to significantly outperform neural operators such as FNO in solving PDEs \cite{gupta2023towards}. However, recent transformer-based FM approaches primarily compare against U-Net on a single family of PDEs \cite{mccabe2024multiple, hao2024dpot, shen2024ups}, overlooking its potential as a foundation model architecture—particularly given the availability of large-scale PDE datasets from diverse systems.

PDE FMs are formulated in various ways with different assumptions on the form of input data and output predictions. Other FMs rely on multiple previous timesteps as input and predict output trajectories, or include temporal information in the input \cite{mccabe2024multiple, herde2024poseidon, hao2024dpot}. SPUS performs the most generalizable task of autoregressively predicting the next timestep given the current timestep. Not only is this problem formulation the most general approach, an operator predicting a single timestep more closely replicates the behavior of numerical solvers, potentially learning the underlying physics \cite{lippe2023pde}. 

This work addresses the following key questions regarding FMs for PDEs:
\begin{itemize}
    \item[(a)] Rather than designing a new and complex architecture, can we utilize a simple, existing one—such as residual U-Net—as an FM for PDEs?
    \item[(b)] Can a lightweight, low-parameter FM achieve state-of-the-art generalization on unseen PDEs?
    \item[(c)] Can pretraining on a set of simpler PDEs but exhibiting diverse physical behaviors (e.g., shocks, shear, vorticity) enable effective transfer to downstream tasks governed by complex PDEs and dominant dynamics, such as vortex evolution from piecewise-constant or shear-layer initial conditions?
    \item[(d)] Can an FM be pretrained to emulate the behavior of numerical solvers by autoregressively predicting the next time step from the current one, thereby potentially learning the underlying physics?
\end{itemize}

Finally, we demonstrate that SPUS, built on a simple residual U-Net and pretrained to emulate the behavior of numerical solvers, achieves state-of-the-art generalization with a lightweight design, transfers knowledge effectively from simpler PDEs to more complex ones, and thereby establishes a path toward efficient, generalizable PDE foundation models.

\section{Preliminaries}
\label{Preliminaries}
PDEs model a wide range of physical phenomena and include equations such as Navier-Stokes, compressible Euler, the Wave equation and others. The general form of a time-dependent PDE is:
\begin{equation}
    \begin{split}
        \delta_t &u(x,t) + L(u, \nabla_x u, \nabla_x^2 u, \ldots) = 0, 
        \\ &\forall x \in D \subset \mathbb{R}^d, 
        \quad t \in (0,T), 
        \quad \mathcal{B}(u) = 0,
        \\ &\forall(x,t) \in \delta D \times (0,T),
        \quad u(x, 0) = u^0(x), 
        \quad x\in D
    \end{split}
\end{equation}
for given boundary conditions $\mathcal{B}$ and initial conditions $u^0$.
Many PDE datasets are discretized in space and time. We denote the discretized spatial state at each timestep as $X_t = \{(x_j,u_j^t) : x_j \in \mathcal{X}\}, t \in [0, 1, \ldots, n]$ where $\mathcal{X}$ is the discretized spatial mesh and $n$ is the number of discretized timesteps. Initial conditions are given by $X_{t=0}$ and each $X_t \in \mathbb{R}^d$ where $d$ is the dimensionality of system variables.

\section{Relevant Work}
\label{Relevant Work}
The closest PDE FMs to ours fall into three distinct formulations. 

\begin{itemize}
    \item[(a)] PDE FMs which take $\{X_{t=[0,m]}\}$ of a PDE trajectory as input and autoregressively predict $\{X_{t=[m+1,n]}\}$; where $m=15$ for MPP \cite{mccabe2024multiple}, and $m=9$ for DPOT \cite{hao2024dpot}. MPP projects normalized field variables from diverse physical systems into a unified latent space and utilizes an axial attention vision transformer-based architecture to perform autoregressive prediction over multiple systems. On the other hand, DPOT injects small-scale noise to $\{X_{t=[0,m]}\}$ and utilize a Fourier attention based transformer architecture to perform autoregressive prediction over multiple systems.

    \item[(b)] PROSE FD \cite{liu2024prose} which takes $\{X_{t=[0,m]}\}$ of a PDE trajectory as input and simultaneously predict $\{X_{t=[m+1,n]}\}$ as a trajectory where $m=9$ and $n=19$. PROSE FD introduces a multimodal transformer framework which takes $\{X_{t=[0,m]}\}$ of a PDE trajectory and mathematical description of the physical behavior as input and performs simultaneous prediction for multi-physics systems.

    \item[(c)] POSEIDON \cite{herde2024poseidon} which takes $(X_{t=0}, \Delta t)$ as input and predicts $\{X_{t=\Delta t}\} \forall \Delta t \in [1, T]$ where $T=14$. POSEIDON proposes a multiscale operator transformer architecture enhanced with time-conditioned normalization to perform prediction on multiple physical systems. Similar to SPUS, POSEIDON uses only a single time step (rather than a trajectory) as input; however, instead of performing autoregressive rollout, it predicts arbitrary future time steps directly. For a dataset with \(n\) time steps, POSEIDON trains on \(O(n^2)\) input-output pairs, whereas our approach is more sample-efficient, requiring only \(O(n)\) sequential pairs.

\end{itemize}

\section{Methods}
\label{Methods}
SPUS is a lightweight, low-parameter residual U-Net architecture designed for modeling PDE dynamics. To enable effective learning within this compact model, we utilize an auto-regressive pretraining scheme.  This method facilitates the efficient modeling of temporal dynamics of PDEs with reasonable accuracy and low computational overhead.
\paragraph{Problem statement}  
Given an initial state $X_{t=0}$ of a trajectory governed by a specific PDE, where $X_t \in \mathbb{R}^d$ represents the system state at time step $t$ with $d$ variables, our objective is to predict the future states $X_{t=1}, X_{t=2}, \dots, X_{t=n}$. 

\paragraph{Auto-regressive pretraining and finetuning}
We formulate the problem as a \emph{first-order Markov process} \cite{pillai2002probability}, in which the evolution of the system depends only on the immediately preceding state. That is, the prediction of $X_{t+1}$ is conditioned solely on $X_t$, satisfying the Markov property: 
\begin{equation}
P(X_{t+1} \mid X_t, X_{t-1}, X_{t-2}, \dots, X_0) = P(X_{t+1} \mid X_t).
\label{eq:mkv}
\end{equation}
This formulation allows the system dynamics to be modeled using an autoregressive framework consistent with the Markov assumption. 

\begin{figure}[ht]
\centering
\includegraphics[width=0.8\linewidth]{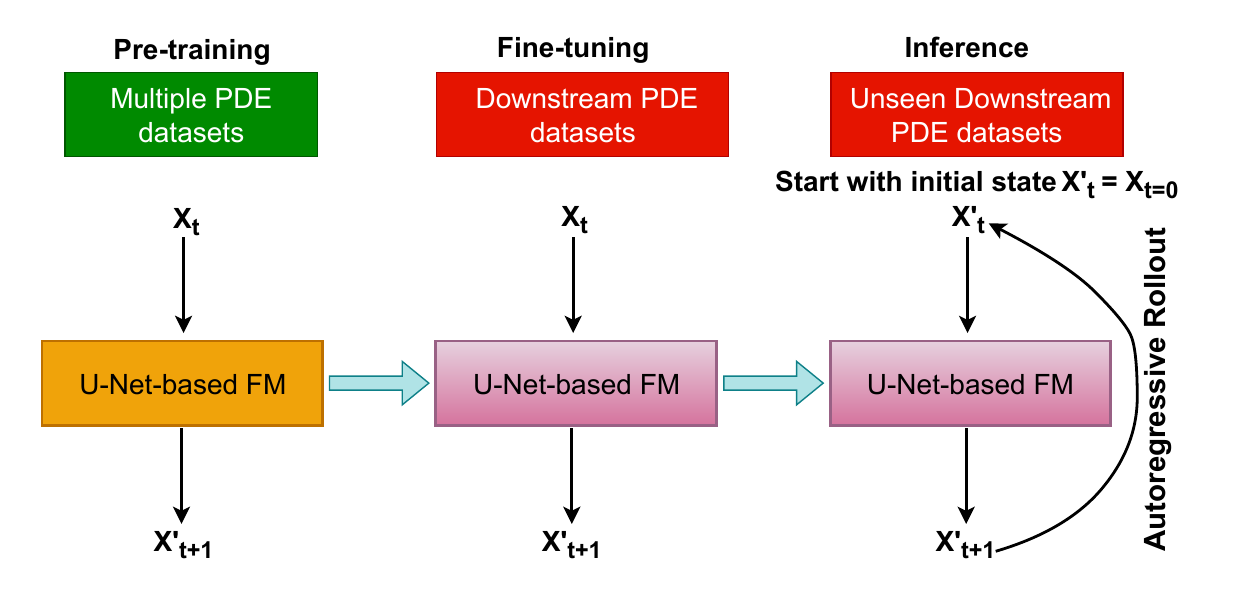}

\caption{Proposed auto-regressive training methodology for the U-Net-based FM. During both pretraining and finetuning, the FM randomly samples a ground truth state $X_t$, where $X_t\in\mathbb{R}^d$ represents the system variables at time step $t$, and learns to predict the next state $X'_{t+1}$. During inference, the full trajectory is predicted autoregressively from the initial condition $X_{t=0}$. The FM takes $X'_t = X_{t=0}$ as input and recursively predicts subsequent states based on its own previous outputs for $t = 1, \ldots, n$, where $n$ is the maximum length of the trajectory to be considered.
}
\label{fig_methodology}
\end{figure}

The proposed auto-regressive training methodology for the U-Net-based FM is illustrated in Figure~\ref{fig_methodology}. During pretraining, the proposed FM takes a randomly sampled ground-truth state $X_t$ from a PDE trajectory in the pretraining dataset and predicts the next state $X'_{t+1}$. More specifically, during pretraining, only ground-truth states are used as inputs; predicted states are not used to generate future predictions. During finetuning, the pretrained model is adapted to a specific downstream PDE using the same input-output structure as in pretraining: the model receives $X_t$ and predicts $X'_{t+1}$. At inference time, however, we provide the model with the initial state $X_{t=0}$ and auto-regressively generate predictions $X'_{t=1}, X'_{t=2}, \dots, X'_{t=n}$, where each prediction $X'_{t+1}$ is based on the previously predicted state $X'_t$ as shown in Figure~\ref{fig_methodology}.

\subsection{Model Architecture}
\label{Model Architecture}
\begin{figure}[ht]
\centering
\includegraphics[width=0.95\linewidth]{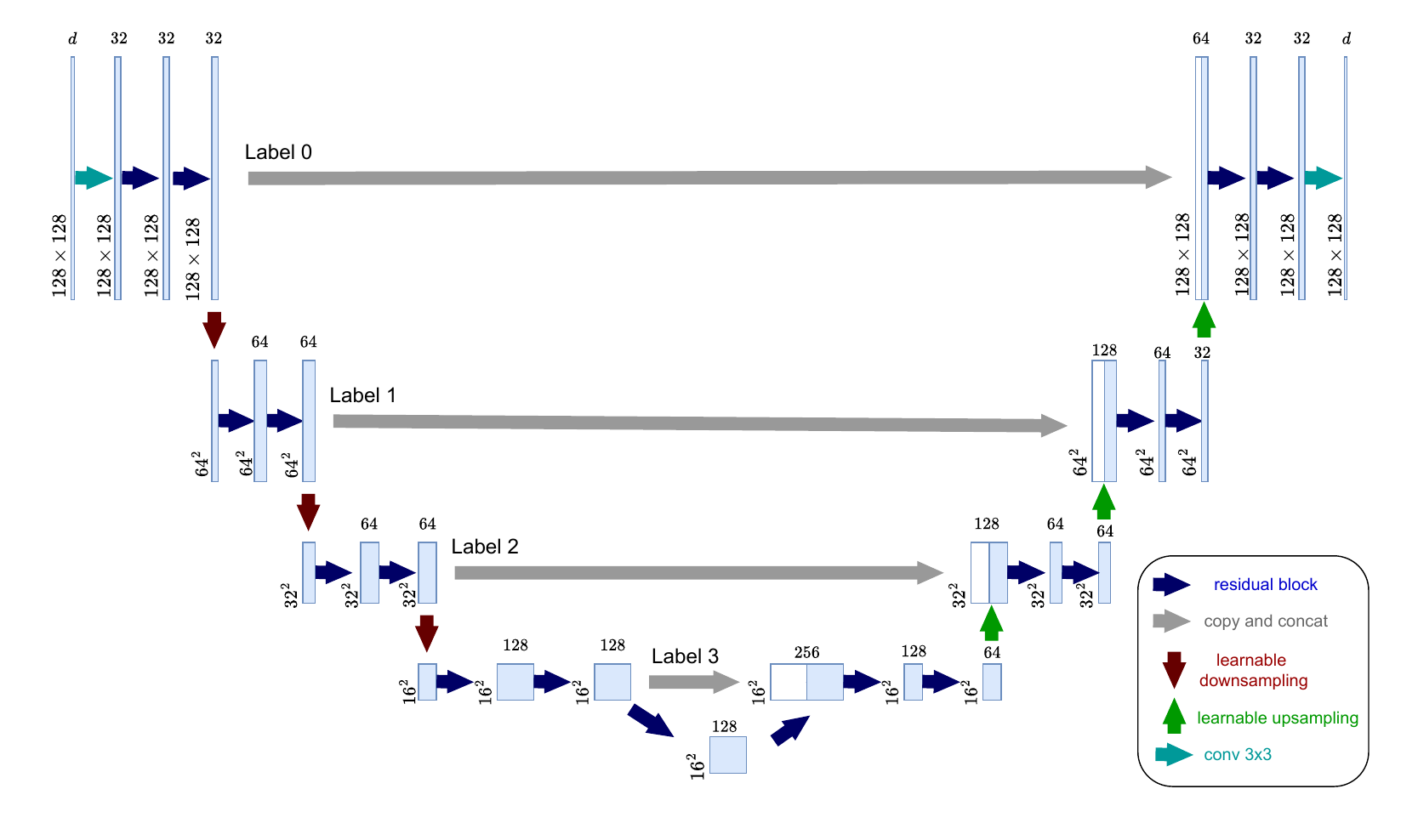}
\caption{Illustration of the residual U-Net based FM architecture for PDEs with 36M parameters. The network takes an input of shape $d\times128\times128$, representing the current time step of a PDE trajectory, and predicts the next time step of the same shape. It employs an encoder–decoder structure with residual blocks, skip connections, and progressive downsampling and upsampling to preserve spatial and contextual information.}

\label{fig_unet}
\end{figure}

Figure \ref{fig_unet} shows the residual U-Net architecture \cite{lan2020real, ronneberger2015u} with 36 million parameters we have utilized for designing the FM for PDEs. The U-Net model takes any current state $X_t$ of shape $d\times128\times128$, where $d$ is the number of system variables and applies a $3\times3$ convolutional layer to project it into a 32-dimensional feature space. The encoder path comprises four hierarchical levels, each of which processes features through two residual blocks. Each residual block includes two $3\times3$ convolutional layers, with batch normalization \cite{bjorck2018understanding} and GELU activation \cite{hendrycks2016gaussian} applied after each convolution, and incorporates a skip connection to preserve feature integrity and support gradient flow. Strided convolution is applied for downsampling at each label of encoder except the last. The encoder processes features through 32-channel blocks at Level 0, increases to 64 channels at Levels 1 and 2, and reaches 128 channels at Level 3. The bottleneck consists of two residual blocks that operate on 128-channel feature maps, effectively capturing high-level representations. The decoder mirrors the encoder with three upsampling stages implemented with transposed convolution layers. At each stage, the upsampled features are concatenated with the corresponding encoder feature maps via skip connections, facilitating the recovery of spatial details. After concatenation, the features are passed through two residual blocks, each followed by batch normalization and GELU activation. The decoder progressively reduces the feature dimensionality from 128 to 64, and subsequently from 64 to 32 across its stages. Finally, a $3\times3$ convolution maps the decoder output back to the number of system variable $d$.

During pretraining, the model was trained for 200 epochs using the Adam optimizer \cite{adam2014method}  with a linear learning rate schedule starting from $10^{-4}$, and a batch size of 10. The learning rate decreased linearly over the course of training. The model achieving the best performance on the evaluation set of the pretraining dataset was saved for downstream use.

\subsection{Pretraining and Finetuning Dataset}
\label{Pretraining and Finetuning}
SPUS is pretrained on a diverse set of PDE types from the \textsc{PDEgym} dataset \cite{herde2024poseidon}, which includes four operators derived from the compressible Euler (CE) equations:
\begin{itemize}
    \item \textbf{CE-RP}, containing trajectories initialized with four-quadrant Riemann problems;
    \item \textbf{CE-CRP}, initialized with multiple curved Riemann problems;
    \item \textbf{CE-KH}, representing shear-driven Kelvin--Helmholtz instabilities; and
    \item \textbf{CE-Gauss}, featuring initial conditions with Gaussian vorticity profiles.
\end{itemize}
Each dataset consists of 10,000 trajectories. Each trajectory has 21 time steps and each time step consists of five physical fields: density $\rho$, horizontal velocity $u$, vertical velocity $v$, pressure $p$, and energy $E$ with spatial grid of resolution $128\times128$.

We fine-tune SPUS on six previously unseen downstream PDEs from the \textsc{PDEgym} dataset, using 128 trajectories for each PDE task. These downstream PDEs include two operators governed by the CE equations, three operators governed by the incompressible Navier-Stokes (NS) equations, and one based on the wave equation:
\begin{itemize}
    \item \textbf{CE-RPUI:} consisting of trajectories initialized with four-quadrant Riemann problems featuring uncertain interfaces;
    \item \textbf{CE-RM:} representing the Richtmyer-Meshkov instability problem;
    \item \textbf{NS-PwC:} initialized from piecewise-constant vorticity fields;
    \item \textbf{NS-SL:} initialized with double shear layer conditions;
    \item \textbf{FNS-KF:} also initialized from piecewise-constant vorticity fields;
    \item \textbf{Wave-Gauss:} containing trajectories initialized as a sum of Gaussians that are propagated by the spatially varying wave speed.
\end{itemize}
Each CE-RPUI and CE-RM trajectory contains 21 time steps. Each time step has five physical fields: density $\rho$, horizontal velocity $u$, vertical velocity $v$, pressure $p$, and energy $E$. On the other hand, each trajectory in the three NS datasets also has 21 time steps but only two physical fields: horizontal velocity $u$, and vertical velocity $v$. For the Wave-Gauss dataset, trajectories have 15 time steps with one physical field, spatially varying wave speed $w$. All fine-tuning datasets share a common spatial resolution of $128\times128$ grid points.

\subsection{Finetuning strategies and Baseline Models}  
In downstream tasks, the number of variables per time step may differ from those used during pretraining. To adapt the pretrained model to downstream tasks with different input and output dimensions than pretraining, we introduce lightweight input and output adapters. Specifically, we use $1 \times 1$ convolutional layers as adapters:
\begin{itemize}
    \item The \textbf{InputAdapter} maps the task-specific input (e.g., 2 fields for NS-SL) to the 5-field format expected by the pretrained SPUS model.
    \item The \textbf{OutputAdapter} maps the model’s 5-field output back to the task-specific output dimensionality (e.g., 2 fields for NS-SL).
\end{itemize}

These adapters are simple, efficient, and allow the pretrained model to be flexibly applied to a variety of downstream tasks without modifying its internal architecture. For each downstream task, we fine-tuned either the pretrained model or the pretrained model with adapters (if the number of the fields differed from five) using 128 trajectories. The model was fine-tuned for 200 epochs using the Adam optimizer  with a linear learning rate schedule starting from $10^{-4}$, and a batch size of 10. The learning rate decreased linearly over the course of training. 

To ensure a fair comparison, we fine-tune two baseline FMs: DPOT ``M'' (122M parameters)~\cite{hao2024dpot} and POSEIDON ``B'' (158M parameters)~\cite{herde2024poseidon}. DPOT was pretrained on 12 PDE datasets governed by the Navier-Stokes, diffusion-reaction, and shallow-water equations, whereas POSEIDON was pretrained on 6 PDE datasets governed by the compressible Euler and Navier-Stokes equations. For both baselines, we adopt the exact hyperparameter settings recommended in their original papers and accompanying code repositories~\cite{hao2024dpot, herde2024poseidon}. All models, including SPUS, are fine-tuned separately on each downstream PDE task using the same set of 128 trajectories for 200 epochs. Performance is evaluated on testing dataset corresponding to each PDE task.

DPOT recommends a context window of 10 timesteps. Accordingly, to predict trajectories from their initial conditions, we follow the same fine-tuning methodology described in~\cite{herde2024poseidon}, padding input sequences with timestep 0 when predicting steps earlier than the 10\textsuperscript{th}. For instance, to predict the state at timestep 4, the input sequence is padded as follows:
\[
[\text{ts}_0, \text{ts}_0, \text{ts}_0, \text{ts}_0, \text{ts}_0, \text{ts}_0, \text{ts}_0, \text{ts}_1, \text{ts}_2, \text{ts}_3].
\]

POSEIDON, on the other hand, is designed to take a single timestep as input, along with the corresponding $\Delta t$, and directly predict any future frame within the trajectory. This allows POSEIDON to predict any timestep (using only the initial timestep as context) without requiring an autoregressive rollout. In practice, such ``direct'' predictions result in higher average accuracy compared to predictions generated via autoregressive rollout. Therefore, we report POSEIDON's performance based on its direct prediction accuracy. 

The comparison of model performance---measured as average mean squared error (MSE) across all predicted timesteps from the initial condition of the trajectories---on six unseen downstream PDE datasets fine-tuned with 128 trajectories is presented in Table~\ref{tab:results}.

\begin{table*}[t]
\centering
\caption{Comparison of model performance (average MSE on all predicted timesteps from initial condition of the trajectories) on six unseen downstream PDE datasets finetuned with $128$ trajectories. Lower is better.}
\begin{tabular}{l c cccccc}
\toprule
Model 
& Params 
& CE-RPUI 
& CE-RM 
& NS-PwC 
& NS-SL 
& FNS-KF 
& Wave-Gauss \\
\midrule

SPUS (Ours)     & \textbf{36M}  & \textbf{0.0054} & \textbf{0.0159} & 0.0048 & \textbf{0.0163} & \textbf{0.0015} &   0.0069     \\
DPOT            & 122M &    0.0570    &   0.0222     &    0.02942    &   0.1461     &   0.0301     &   0.0107     \\
POSEIDON   & 158M &   0.0085     &   0.4181     &    \textbf{0.0004}    &    \textbf{0.0163}   &   0.0017    &    \textbf{0.0068}    \\
\bottomrule
\label{tab:results}
\end{tabular}
\end{table*}

\begin{figure*}[ht]
\centering
\includegraphics[width=\linewidth]{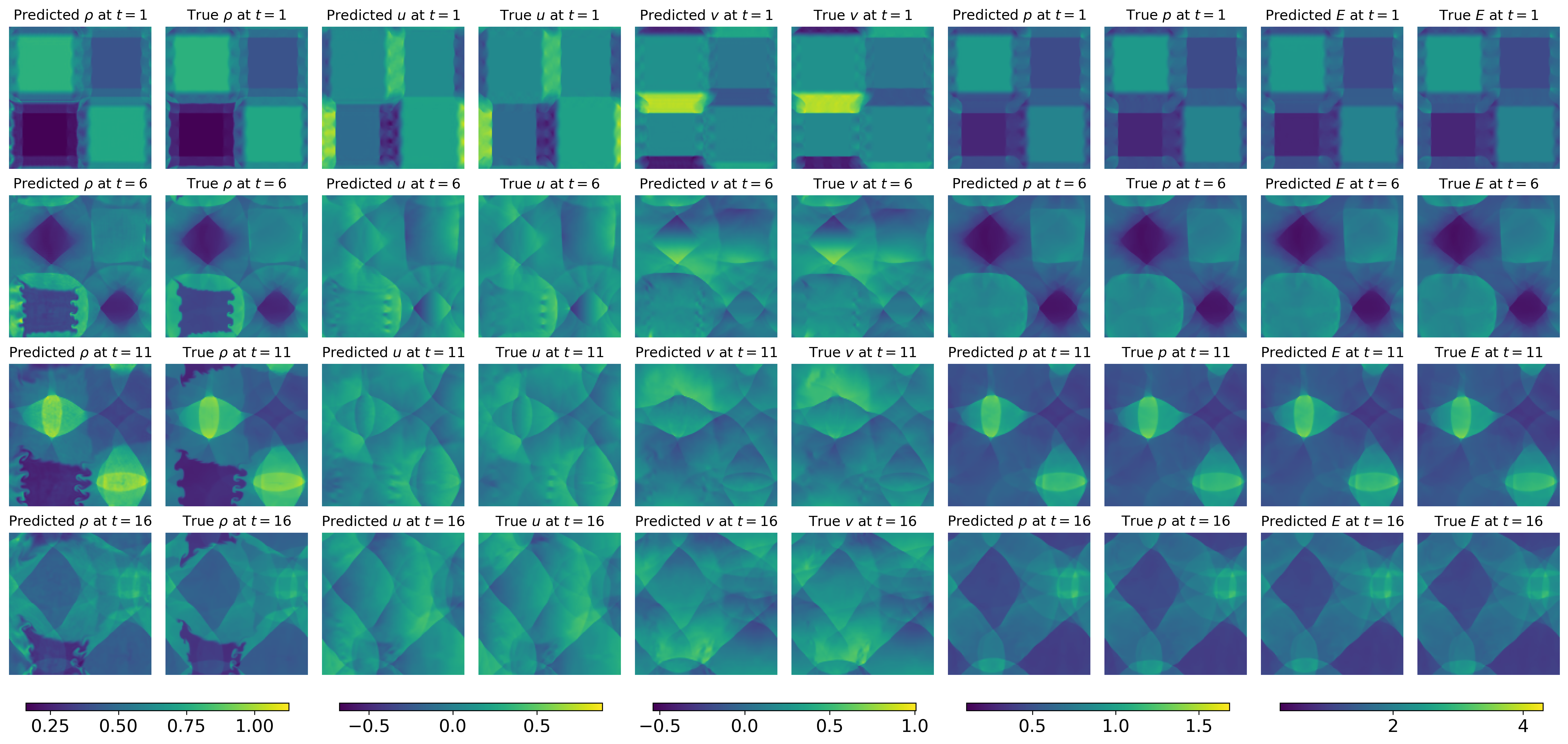}
\caption{Autoregressive trajectory prediction by SPUS from the initial condition of a randomly selected trajectory in the CE-RPUI testing dataset (240 test trajectories). The figure shows example results at time steps $t = 1, 6, 11, 16$ for five system variables: density $\rho$, horizontal velocity $u$, vertical velocity $v$, pressure $p$, and energy $E$. SPUS takes the initial condition $X'_t = X_{t=0}$ as input and recursively predicts subsequent states based on its own previous outputs for $t = 1, \ldots, 20$, as described in Figure~\ref{fig_methodology} (inference step). As shown, the predicted variables closely match the ground truth at each time step.}
\label{fig_CE-RPUI}
\end{figure*}

\begin{figure*}[ht]
\centering
\includegraphics[width=\linewidth]{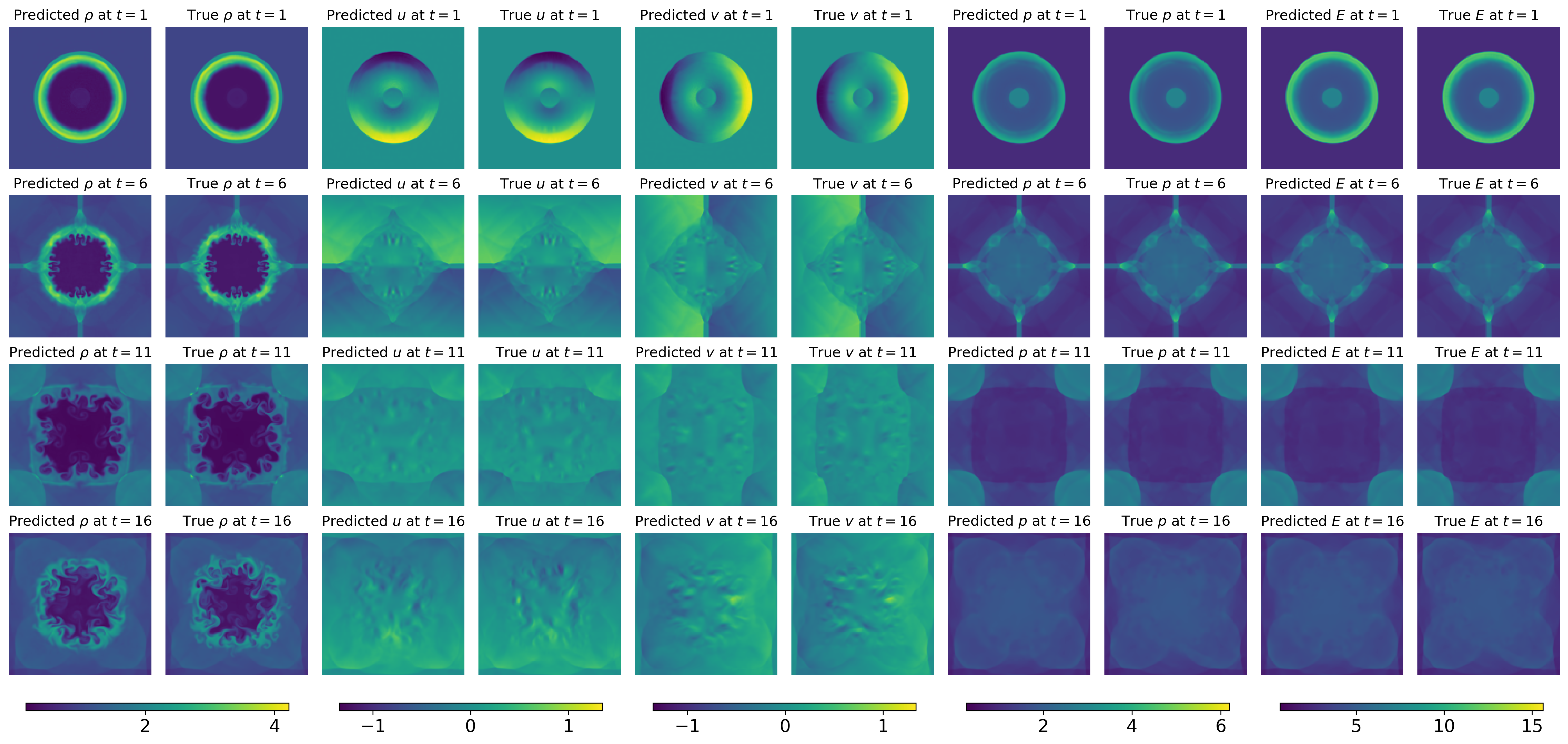}
\caption{Autoregressive trajectory prediction by SPUS from the initial condition of a randomly selected trajectory in the CE-RM testing dataset (130 test trajectories). The figure shows example results at time steps $t = 1,\ 6,\ 11,\ 16$ for five system variables: density $\rho$, horizontal velocity $u$, vertical velocity $v$, pressure $p$, and energy $E$. SPUS takes the initial condition $X'_t = X_{t=0}$ as input and recursively predicts subsequent states based on its own previous outputs for $t = 1,\ldots,20$, as described in Figure~\ref{fig_methodology} (inference step). As shown, the predicted variables closely match the ground truth at each time step, although the deviation between prediction and ground truth increases more noticeably over time for CE-RM compared to CE-RPUI due to its more complex dynamics.
}
\label{fig_CE-RM}
\end{figure*}

\section{Experiments}
\label{Experiments}
\paragraph{Is SPUS an effective lightweight PDE emulator? Does SPUS with only 36 million parameters generalize as accurately as larger models?}To address these questions, we design and evaluate the following two experiments.
\paragraph{(A). Does SPUS generalize to unseen systems governed by the compressible Euler (CE) equations, consistent with its pretraining?}
We investigate whether SPUS can generalize to previously unseen physical systems that are governed by CE equations, consistent with its pretraining. To evaluate this, we fine-tune the pretrained SPUS model on the CE-RPUI dataset. While this dataset adheres to the same underlying physical laws, its distribution of initial conditions differs from those seen during pretraining, presenting a clear out-of-distribution (OOD) generalization challenge \cite{herde2024poseidon}. As shown in Table~\ref{tab:results}, SPUS achieves strong performance in autoregressively predicting full trajectories from initial conditions, despite having only 36 million parameters. Notably, it outperforms both the substantially larger POSEIDON model (158 million parameters) and the DPOT model (122 million parameters) in terms of average mean squared error (MSE) across 240 test trajectories. A randomly selected trajectory prediction from the CE-RPUI test set is shown in Figure~\ref{fig_CE-RPUI}, where the SPUS predictions closely match the ground truth at each time step. These results demonstrate the effectiveness and computational efficiency of the lightweight SPUS model relative to significantly larger architectures.

We also fine-tune the pretrained SPUS model on the CE-RM dataset, which exhibits significantly more complex dynamics compared to CE-RPUI. SPUS demonstrates strong generalization capability in predicting entire trajectories from initial conditions, as illustrated in Figure~\ref{fig_CE-RM}. Furthermore, as shown in Table~\ref{tab:results}, SPUS achieves a lower average MSE across 130 test trajectories compared to both the POSEIDON and DPOT models, despite their substantially larger parameter counts.

\paragraph{(B). Does SPUS generalize to systems governed by equations different from those used in pretraining?}
We investigate the ability of SPUS to generalize to previously unseen physical systems governed by equations different from those used during pretraining. Specifically, we fine-tune the pretrained SPUS model on three datasets governed by incompressible NS equations that were not part of the pretraining data: NS-PwC, NS-SL, and FNS-KF. As shown in Table~\ref{tab:results}, despite not being exposed to incompressible NS dynamics during pretraining, surprisingly, SPUS achieves superior time-step prediction performance compared to DPOT across all three datasets—even though DPOT was pretrained on operators of both compressible and incompressible NS equations. For the POSEIDON model, whose pretraining data includes two operators governed by NS equations, SPUS outperforms it on FNS-KF, matches its performance on NS-SL, and is  outperformed on NS-PwC, as summarized in Table~\ref{tab:results}. These results demonstrate the strong generalization capability of SPUS to new physical regimes outside its pretraining distribution and highlight its effective transferability to downstream tasks governed by equations different from those seen during pretraining. Randomly selected trajectory predictions from the test datasets of NS-PwC, NS-SL, and FNS-KF are shown in Figure~\ref{fig:NS-PwC-SL} and Figure~\ref{fig:FNS-KF} (in Appendix). As observed, SPUS demonstrates strong generalization performance on each of the NS dataset; however, the predicted variables gradually deviate from the ground truth over time. We also fine-tuned SPUS on the Wave-Gauss dataset, which is governed by the wave equation. As shown in Table~\ref{tab:results}, SPUS outperforms DPOT and is narrowly outperformed by POSEIDON for 240 Wave-Gauss testing trajectories.

\begin{figure}[ht]
\centering
\begin{subfigure}[b]{0.49\linewidth}
    \includegraphics[width=\linewidth]{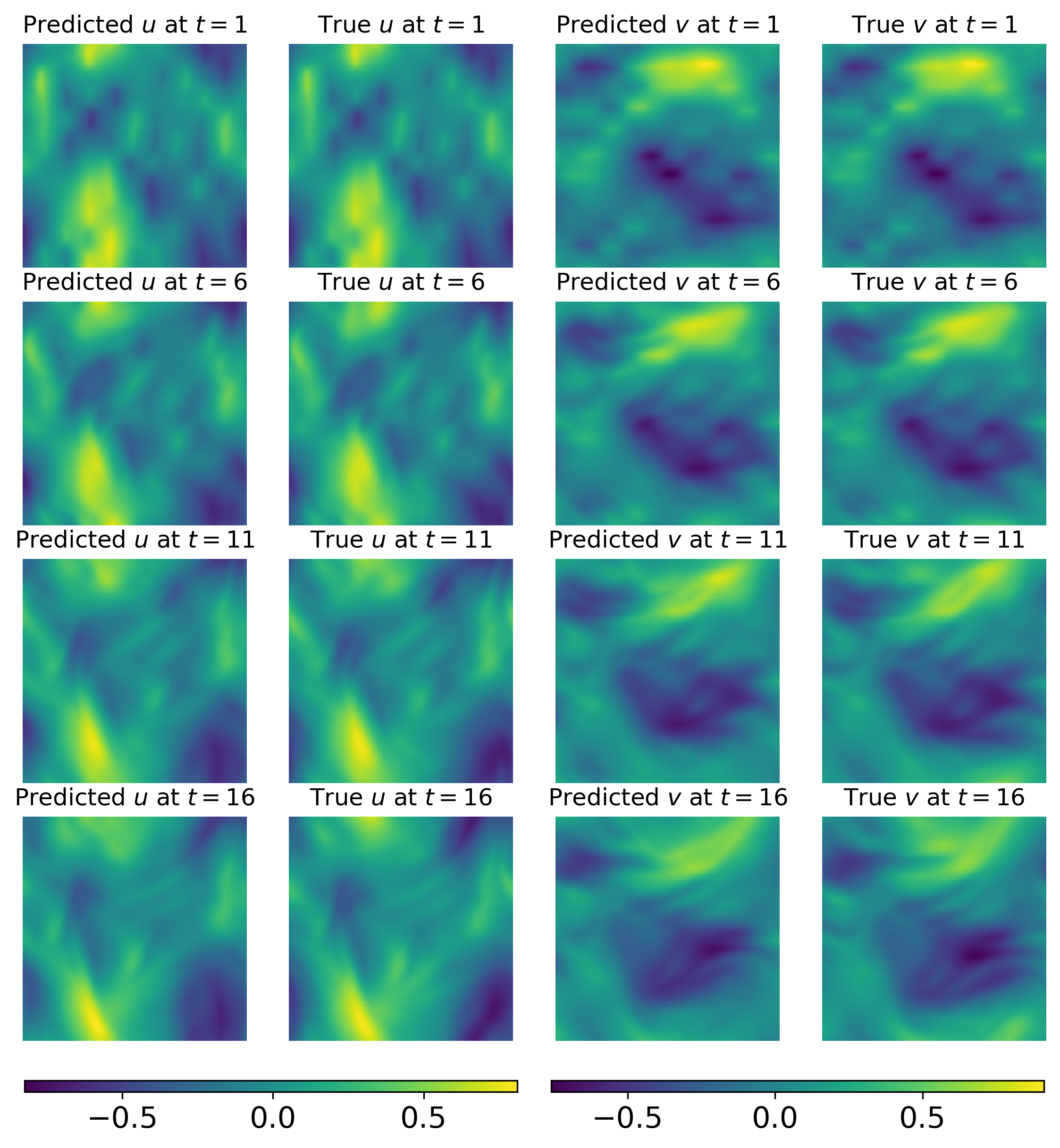}
    \caption{Autoregressive trajectory prediction from initial state for NS-PwC}
    \label{fig:subfig_a}
\end{subfigure}
\hfill
\begin{subfigure}[b]{0.49\linewidth}
    \includegraphics[width=\linewidth]{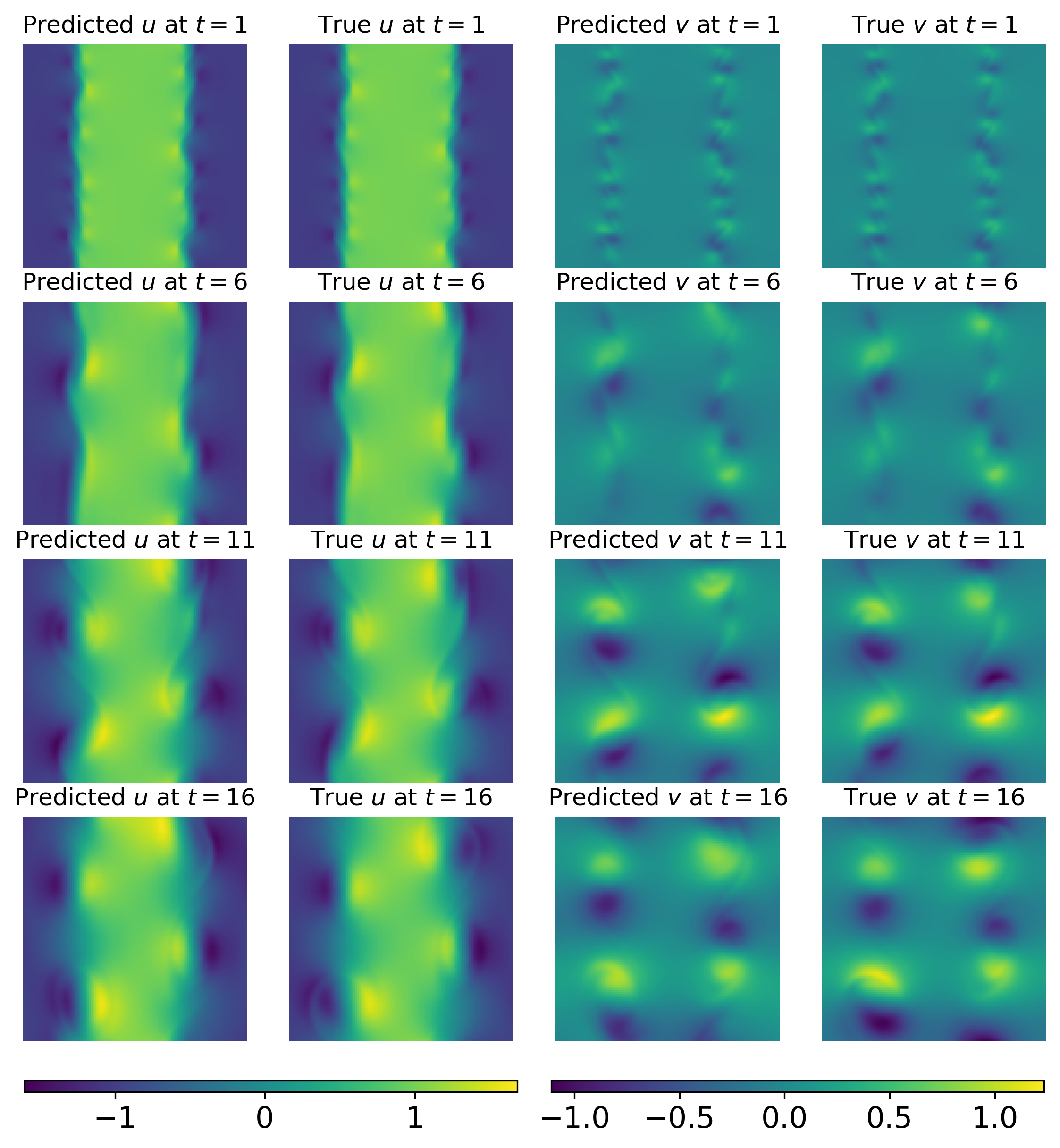}
    \caption{Autoregressive trajectory prediction from initial state for NS-SL}
    \label{fig:subfig_b}
\end{subfigure}
\caption{Autoregressive trajectory prediction by SPUS from the initial condition of a randomly selected trajectory in the NS-PwC and NS-SL testing datasets (each with 240 test trajectories). The figure shows example results at time steps $t = 1,\ 6,\ 11,\ 16$ for two system variables: horizontal velocity $u$, vertical velocity $v$. As shown, despite not being exposed to incompressible NS dynamics during pretraining, the predicted variables by SPUS closely match the ground truth at each time step. The deviation between prediction and ground truth (GT) increases more noticeably over time for NS-PwC compared to NS-SL.}
\label{fig:NS-PwC-SL}
\end{figure}

\paragraph{Does SPUS show scalability with dataset size?}
Table~\ref{tab:datasize} reports the average MSE of predicting the time steps of entire trajectories from the initial condition of the trajectories across six downstream PDEs using SPUS, finetuned with 32, 128, and 256 trajectories. As shown, increasing the fine-tuning set size reduces the MSE across the six downstream PDEs, demonstrating that SPUS scales favorably with additional data.

\paragraph{Summaries of the experiments} 
Based on the above experiments, we observe that SPUS, built on a residual U-Net architecture with only 36 million parameters, achieves state-of-the-art generalization on downstream tasks, outperforming significantly larger models such as POSEIDON (158 million parameters) and DPOT (122 million parameters). These results highlight that a simple, well-established architecture—specifically, a residual U-Net—can be effectively leveraged as a foundation model (FM) for PDEs. Despite its architectural simplicity and relatively small parameter count, SPUS is capable of capturing complex dynamics and performs competitively with more sophisticated, larger models. We also observe that SPUS, when pretrained on a diverse set of simpler PDEs (such as CE), demonstrates strong performance on complex downstream PDEs (such as NS). This indicates the effective transferability of SPUS across distinct physical regimes, despite differences in the underlying governing equations. Furthermore, this suggests that even when the pretraining data are derived from PDEs governed by simple CE equations, a sufficiently diverse pretraining dataset—spanning variations in initial and boundary conditions, domain geometries, and external forcing—can enable the FM to generalize effectively. Moreover, SPUS is pretrained to emulate the behavior of numerical solvers by autoregressively predicting the next time step from the current one. The results on downstream tasks for SPUS suggest that this pretraining strategy helps the FM learn the underlying physics of PDEs, enabling more accurate and physically consistent predictions. 

Additional results on the performance evaluation of SPUS on downstream tasks, as well as visual comparisons of SPUS’s performance on trajectory prediction from initial conditions with larger models (POSEIDON and DPOT), are presented in Appendices~\ref{A1}–\ref{A3}.

\begin{table*}[t]
\centering
\caption{Evaluation (average MSE on all predicted timesteps from initial condition) of SPUS on downstream datasets under different numbers of finetuned trajectories}

\label{tab:datasize}
\begin{tabular}{|c|ccc|}
\hline
\textbf{Downstream Dataset} & \multicolumn{3}{c|}{\textbf{Number of Trajectories}} \\
\cline{2-4}
& \textbf{32} & \textbf{128} & \textbf{256} \\
\hline
CE-RPUI    & 0.0057 & 0.0054 & 0.0041    \\
CE-RM      & 0.0246 & 0.0159 & 0.0130    \\
NS-PwC     & 0.0076 & 0.0048 & 0.0025    \\
NS-SL      & 0.0286 & 0.0163 & 0.0004    \\
FNS-KF     & 0.0098 & 0.0015 & 0.0012    \\
Wave-Gauss & 0.0097 & 0.0069 & 0.0068    \\
\hline
\end{tabular}
\end{table*}

\section{Conclusions}
\label{Conclusions}
We propose SPUS, a compact and lightweight FM for PDEs, capable of handling a broad range of physical systems. The model is based on a simple residual U-Net architecture and is trained using a straightforward autoregressive pretraining strategy. Despite its relatively small size—only 36 million parameters—SPUS demonstrates strong generalization capabilities across six diverse downstream PDE tasks. SPUS consistently outperforms the significantly larger DPOT model across all downstream datasets. When compared to the POSEIDON model, which also has substantially more parameters, SPUS achieves superior performance on three datasets, matches performance on one, is narrowly outperformed on another (MSE: 0.0069 vs. 0.0068), and is outperformed on one task. These results establish SPUS as a highly parameter-efficient foundation model, capable of solving a wide range of complex PDE systems with competitive accuracy. Furthermore, we demonstrate that pretraining SPUS on simpler PDEs (such as CE) with autoregressive training to emulate a numerical solver enables effective transfer to more complex PDEs (such as NS), reducing the amount of data required for finetuning even when the downstream task involves more complex dynamics than those seen during pretraining.

\section*{Reproducibility statement}
All datasets used in this work are publicly available. The code will be released at the time of publication.

\section*{Acknowledgments}
Research presented in this article was supported by the Laboratory Directed Research and Development program of Los Alamos National Laboratory under project number 20250637DI. This research used resources provided by the Los Alamos National Laboratory Institutional Computing Program, which is supported by the U.S. Department of Energy National Nuclear Security Administration under Contract No. 89233218CNA000001.

\bibliographystyle{plainurl}

\bibliography{main}

\clearpage
\appendix

\renewcommand{\thefigure}{A.\arabic{figure}}
\setcounter{figure}{0} %

\section{APPENDIX}
\subsection{Error Growth over Time}
\label{A1}
\begin{figure*}[ht]
\centering
\begin{subfigure}[b]{0.49\linewidth}
    \includegraphics[width=\linewidth]{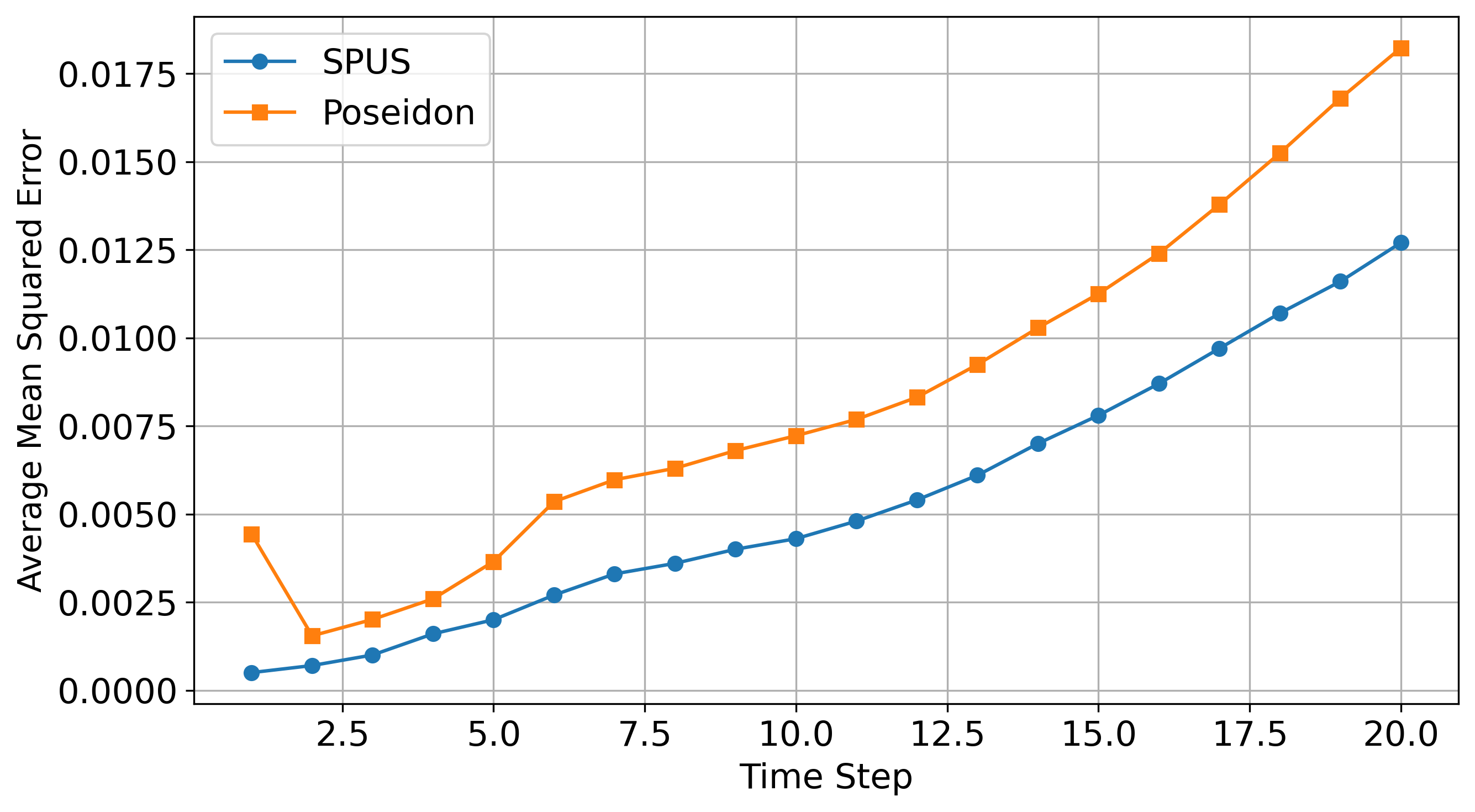}
    \caption{Prediction error over time for CE-RPUI}
    \label{fig:subfig_aa}
\end{subfigure}
\hfill
\begin{subfigure}[b]{0.49\linewidth}
    \includegraphics[width=\linewidth]{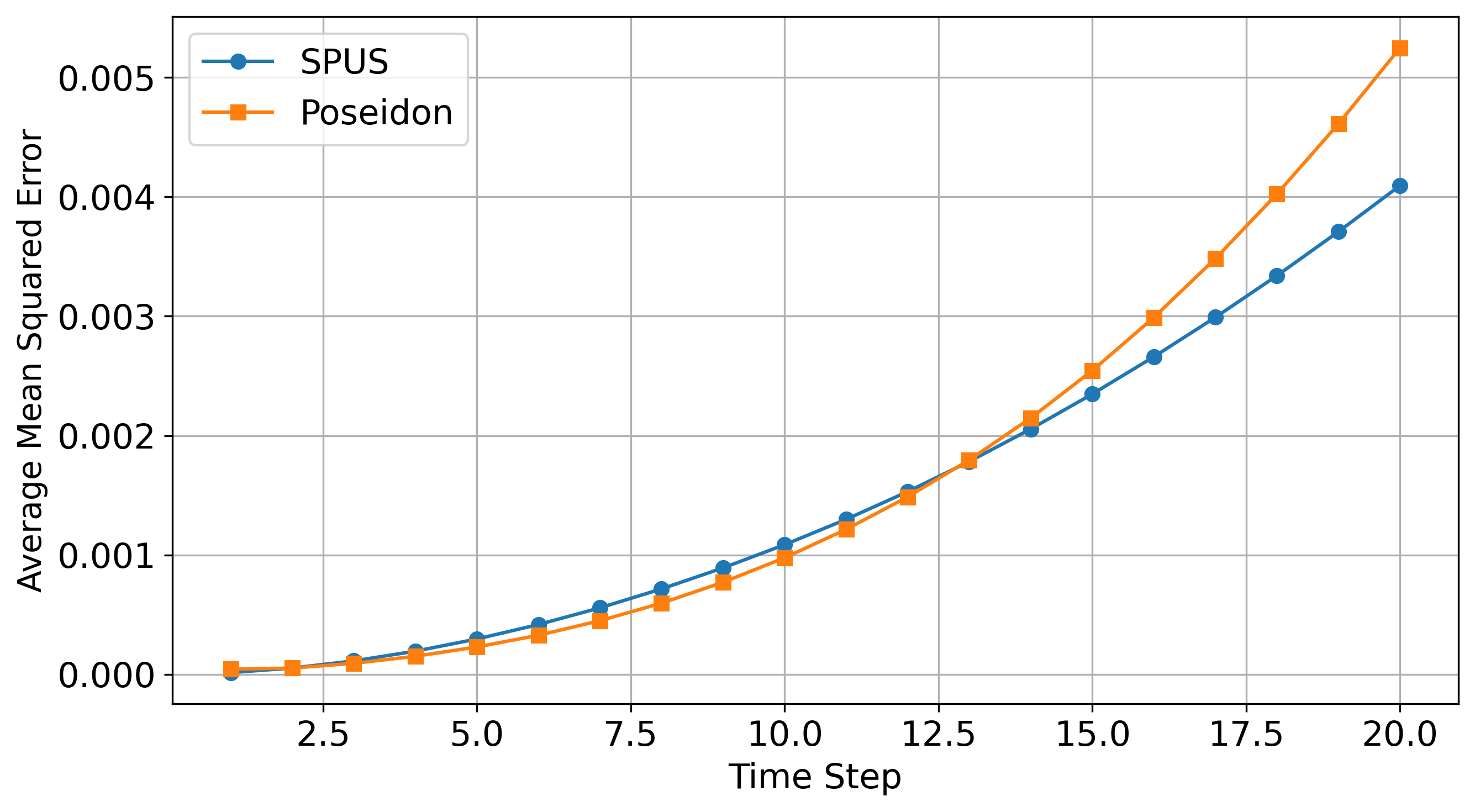}
    \caption{Prediction error over time for FNS-KF}
    \label{fig:subfig_bb}
\end{subfigure}
\caption{Average MSE
of trajectory predictions over time for the CE-RPUI and
FNS-KF test datasets. SPUS (36M) shows approximately linear error growth similar to POSEIDON (158M). It consistently outperforms POSEIDON  at all time steps on CE-RPUI and surpasses POSEIDON at later steps on FNS-KF, highlighting its efficiency and long-term accuracy.}
\label{fig:Prediction-error}
\end{figure*}

Figure~\ref{fig:Prediction-error} presents the average mean squared error (MSE) of trajectory predictions over time for the CE-RPUI and FNS-KF test datasets. As shown, the SPUS model—with only 36 million parameters—exhibits an approximately linear increase in prediction error over time, a behavior consistently observed across our downstream datasets. POSEIDON, a larger FM with 158 million parameters, demonstrates a similar error growth pattern. For the CE-RPUI dataset, SPUS consistently outperforms POSEIDON at all time steps. In the case of FNS-KF, SPUS achieves comparable performance to POSEIDON at early time steps and surpasses it at later time steps. These results highlight the potential of SPUS to deliver accurate long-term predictions despite having significantly fewer parameters.

\subsection{Performance Evaluation of SPUS on the FNS-KF and Wave-Gauss Datasets}
\label{A2}

\begin{figure}[ht]
\centering
\includegraphics[width=\columnwidth]{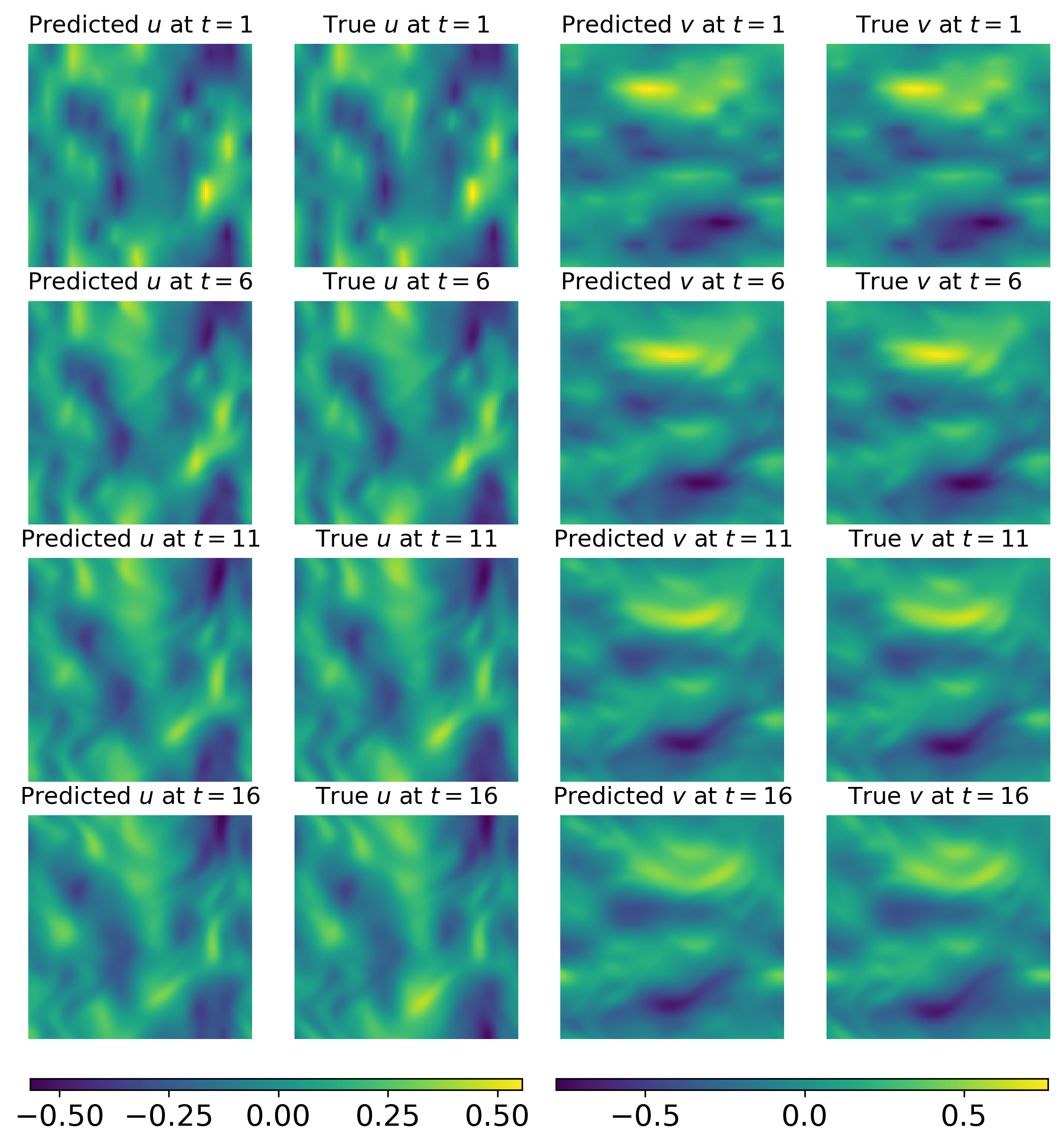}
\caption{Autoregressive trajectory prediction by SPUS from the initial condition of a randomly selected trajectory in the FNS-KF testing dataset (240 trajectories). The figure shows example results at time steps $t = 1, 6, 11, 16$ for two system variables: horizontal velocity $u$, vertical velocity $v$. SPUS takes the initial condition $X'_t = X_{t=0}$ as input and recursively predicts subsequent states based on its own previous outputs for $t = 1, \ldots, 20$, as described in Figure~\ref{fig_methodology} (inference step). As shown, the predicted variables closely match the ground truth at each time step.}
\label{fig:FNS-KF}
\end{figure}

Figure~\ref{fig:FNS-KF} presents randomly selected trajectory prediction from the FNS-KF test dataset. Notably, although SPUS was not pretrained on incompressible Navier–Stokes dynamics, its predictions closely follow the ground truth variables at each time step, demonstrating robust generalization to the FNS-KF system.

\begin{figure}[ht]
\centering
\includegraphics[width=0.7\columnwidth]{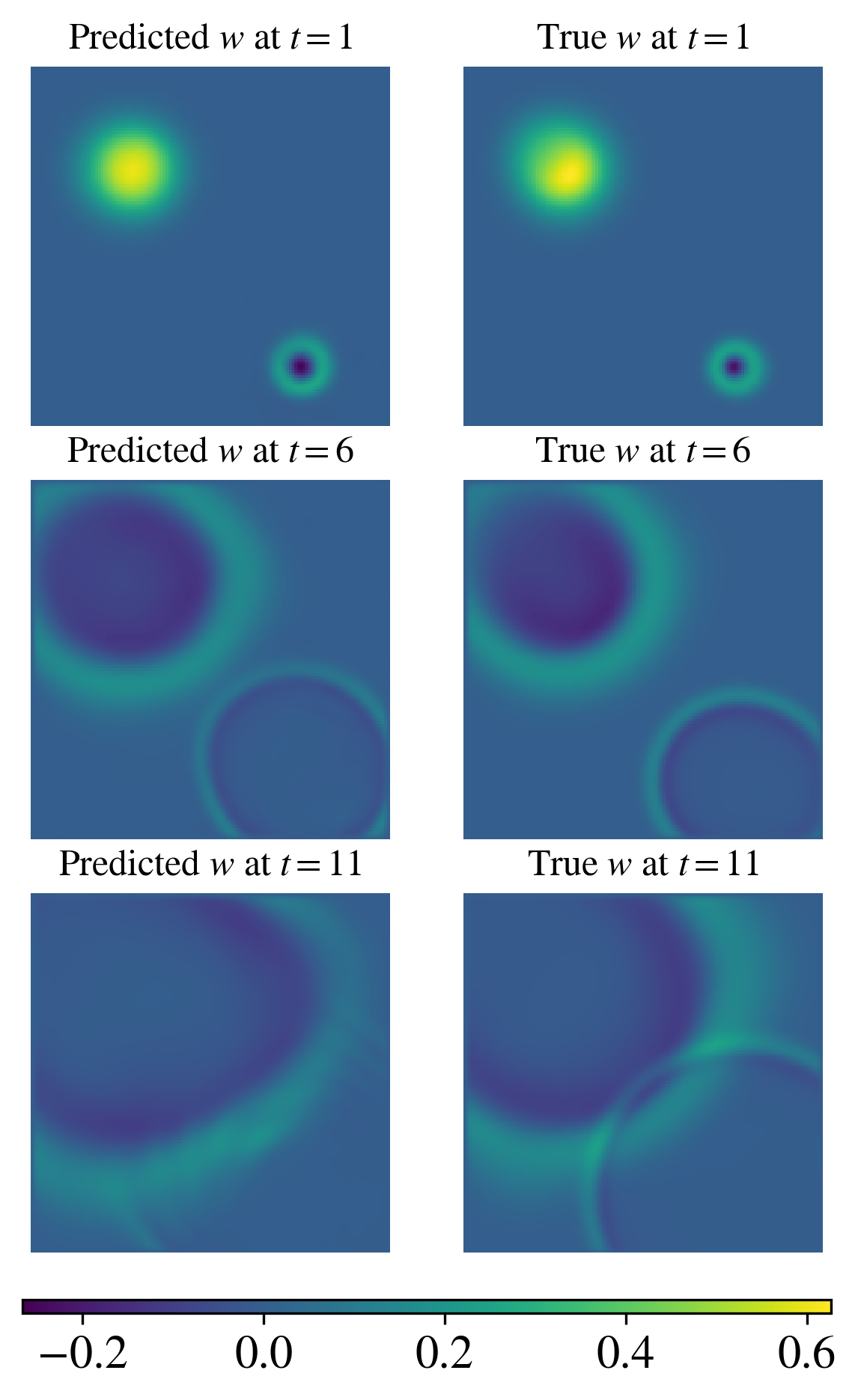}
\caption{Autoregressive trajectory prediction by SPUS from the initial condition of a randomly selected trajectory in the Wave-Gauss testing dataset (240 trajectories). The figure shows example results at time steps $t = 1, 6, 11$ for one system variable: wave speed $u$. SPUS takes the initial condition $X'_t = X_{t=0}$ as input and recursively predicts subsequent states based on its own previous outputs for $t = 1, \ldots, 14$, as described in Figure~\ref{fig_methodology} (inference step). As shown, the predicted variables deviates very quickly from ground truth at each time step for Wave-Gauss compared to other downstream tasks.}

\label{fig_Wave-Gauss}
\end{figure}

Figure~\ref{fig_Wave-Gauss} illustrates a representative trajectory prediction from the Wave-Gauss test set, demonstrating SPUS's ability to generalize to the Wave-Gauss system. However, in comparison to the Navier–Stokes downstream tasks, the deviation between the predicted variables and the ground truth increases more noticeably over time for Wave-Gauss.

\subsection{Visualization of SPUS Performance on Trajectory Prediction Compared with POSEIDON and DPOT}
\label{A3}
\begin{figure*}[ht]
\centering
\includegraphics[width=\textwidth]{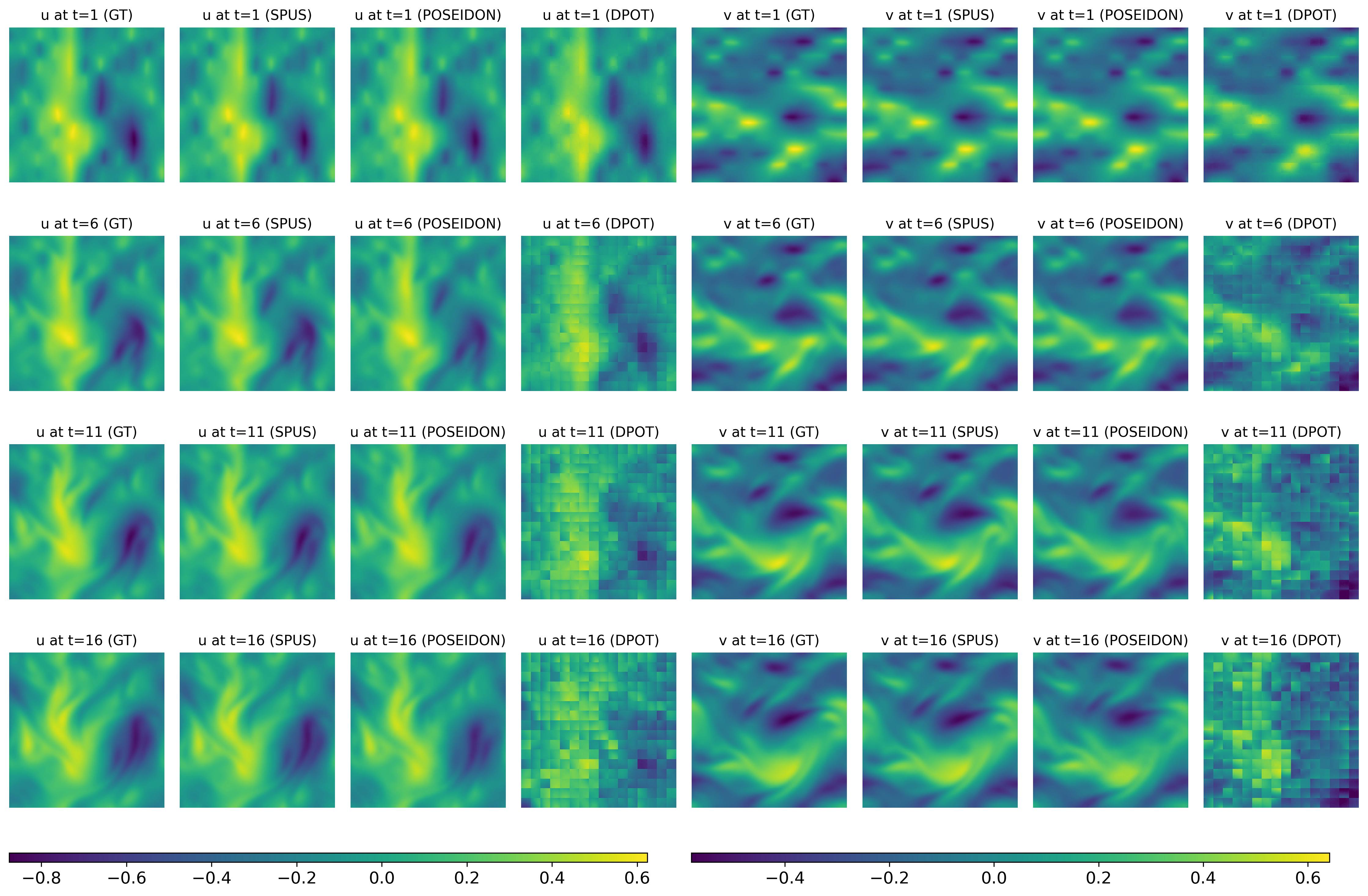}

\caption{A random trajectory predictions for FNS-KF made by SPUS (36M), POSEIDON (158M), and DPOT (122M). The figure shows example results at time steps $t = 1, 6, 11, 16$ for two system variables: horizontal velocity $u$, vertical velocity $v$.}
\label{fig_comparison_FNS_KF}
\end{figure*}

Figure \ref{fig_comparison_FNS_KF} shows a random trajectory predictions for FNS-KF made by SPUS (36M), POSEIDON (158M), and DPOT (122M). Each model is finetuned with 128 trajectories.

\begin{figure*}[ht]
\centering
\includegraphics[width=\textwidth]{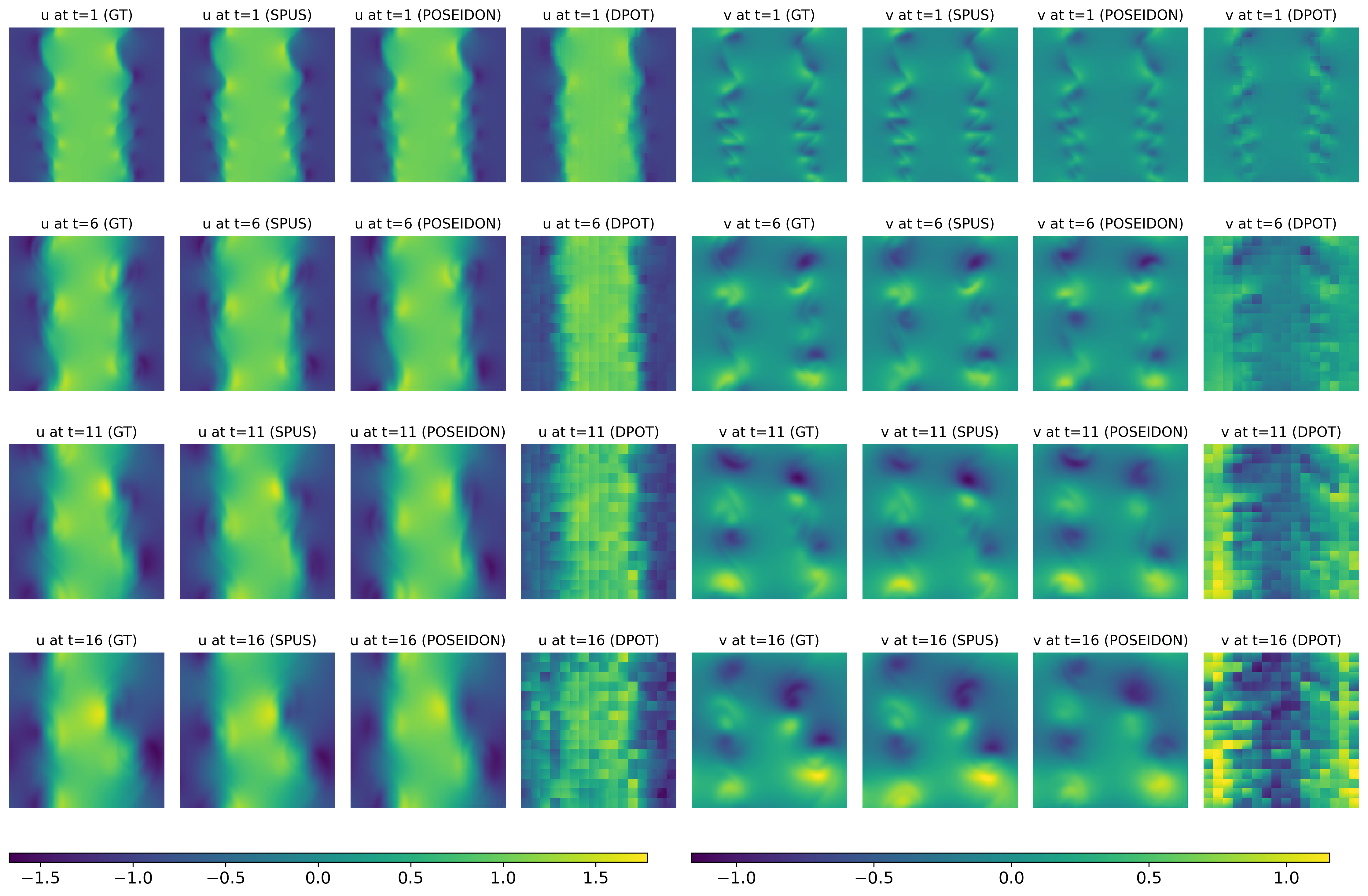}
\caption{A random trajectory predictions for NS-SL made by SPUS (36M), POSEIDON (158M), and DPOT (122M). The figure shows example results at time steps $t = 1, 6, 11, 16$ for two system variables: horizontal velocity $u$, vertical velocity $v$.}
\label{fig_comparison_NS_SL}
\end{figure*}

Figure \ref{fig_comparison_NS_SL} shows a random trajectory predictions for NS-SL made by SPUS (36M), POSEIDON (158M), and DPOT (122M). Each model is finetuned with 128 trajectories.

\end{document}